\documentclass[a4paper]{article}

\usepackage{INTERSPEECH2021}
\usepackage{hyperref}

\def\L{{\cal L}}
\hypersetup{colorlinks=true,citecolor=blue,linkcolor=blue,linktocpage=true}
\newcommand\rpang[1]{}

\title{
Unsupervised Learning of Disentangled Speech \\Content and Style Representation
}
\name{Andros Tjandra$^{1,*}$, Ruoming Pang$^{2}$, Yu Zhang$^{2}$, Shigeki Karita$^{2}$ \thanks{*This work was done while the first author was a research intern at
Google Brain.}}

\address{$^{1}$Nara Institute of Science and Technology, Japan \quad $^{2}$Google LLC}
\email{andros.tjandra@gmail.com, \{rpang,ngyuzh,karita\}@google.com}

\begin{document}

\maketitle
\begin{abstract}
Speech is influenced by a number of underlying factors, which can be broadly categorized into linguistic contents and speaking styles. However, collecting the labeled data that annotates both content and style is an expensive and time-consuming task. Here, we present an approach for unsupervised learning of speech representation disentangling contents and styles. Our model consists of: (1) a local encoder that captures per-frame information; (2) a global encoder that captures per-utterance information; and (3) a conditional decoder that reconstructs speech given local and global latent variables. Our experiments show that (1) the local latent variables encode speech contents, as reconstructed speech can be recognized by ASR with low word error rates (WER), even with a different global encoding; (2) the global latent variables encode speaker style, as reconstructed speech shares speaker identity with the source utterance of the global encoding. Additionally, we demonstrate a useful application from our pre-trained model, where we can train a speaker recognition model from the global latent variables and achieve high accuracy by fine-tuning with as few data as one label per speaker.
\end{abstract}
\noindent\textbf{Index Terms}: Unsupervised, Disentanglement, Speech, VQVAE

\section{Introduction}

% Unsupervised representation learning allows models to learn high-level latent representations from unlabeled data.

% Models pretrained from unsupervised data can be fine-tuned with a small amount of labeled data.
%
% Deep probabilistic generative models presents a powerful approach to learn representations by modeling the data generation process.

% Variational AutoEncoders~\cite{Kingma2014} (VAE) are one of the popular approaches to representation learning by modeling the latent features in a unit Gaussian space.
% Vector-Quantized VAE~\cite{van2017neural} (VQVAE) is a method to learn discrete representations from data.

%
A speech signal is a one-dimensional time-domain representation where each value represents the sound intensity across different time-step. Behind its simple representation, there is a lot of high-level information such as phoneme, tone, emotion, gender, and speaker are mixed underneath.
Learning disentangled latent representations from speech has a wide set of applications in generative tasks, including speech synthesis, data augmentation, voice transfer, and speech compression. Downstream tasks such as speech recognition \cite{schneider2019wav2vec} and speaker classification \cite{oord2019representation} can also benefit from such learned representations.

Because of the cost, complexity, and privacy concerns around collecting labeled speech data, there has been a lot of interest in unsupervised representation learning for speech. Of particular interest is to learn representations for speech styles from unsupervised data due to the difficulty in describing prosody with human labels.

Some previous works aim to learn \textit{global} representations from entire speech sequences. Global style tokens~\cite{DBLP:conf/icml/WangSZRBSXJRS18} learn a dictionary of embeddings from speech without prosody labels. As another example, Hsu et al. \cite{DBLP:conf/iclr/HsuZWZWWCJCSNP19} model disentangled speech styles with a hierarchy of variational autoencoder (VAE)~\cite{Kingma2014}. Hu et al. \cite{hu2020unsupervised} proposed a content and style separation model by pre-training on a single-speaker dataset with text transcription and minimizing mutual information (MI) between the content and style representation. Unfortunately, all of those approaches required datasets with text labels, either in the training or inference stage.

Other works try to learn fine-grain \textit{localized} representations of speech.
\cite{oord2019representation,Schneider2019,baevski2020wav2vec} apply self-supervised learning to unlabeled speech data and extract localized latent representations that can be fine-tuned for speech recognition. Those models produce high-quality content representation over time-step, but the global representation is not captured or disentangled by the model itself.

Factorized Hierarchical Variational Autoencoder (FHVAE)~\cite{FHVAE} learns a sequence of high-level and low-level features by utilizing two-level hierarchical Gaussian VAE. However, the Gaussian VAE-based model has drawbacks such as the evidence lower-bound (ELBO) trade-off between model generation quality and enforcing simple posteriors \cite{yacoby2020failure}. There are also several works based on the generative adversarial network (GAN) model \cite{goodfellow2014gan}. For example \cite{kaneko2018cyclegan, kaneko2019cyclegan} used CycleGAN \cite{zhu2017unpaired} to perform voice conversion without any paired data. While GAN-based models are powerful, they can hard to train and require careful hyperparameter tuning as well \cite{mescheder2018training, lucic2018gans}. \cite{8822475, vqvae2019tjandra,vqvae2020tjandra} leverages vector-quantized VAE~\cite{van2017neural} (VQVAE) to learn a discrete sequence representation of speech. Still, both approaches require the speaker information given during the training stage. 

We propose a simple framework to learn both global and localized representation of speech. In order to disentangle the content and style representations, we apply (1) a local encoder with a VQ layer to learn a discrete per-timestep representation of the speech that captures the linguistic contents and (2) a global VAE to extract per-utterance representations to reflect the speech styles. We further disentangle the local and global representations with a mutual information loss~\cite{pmlr-v80-belghazi18a}. We evaluate the quality of linguistic and style representations by running speech and speaker recognition models on the reconstructed speech. We also show that the global representation captures the speaker information well enough that we can obtain a speaker classification model by training a linear projection layer on top of the global representation with only one example per speaker.

\section{Model architecture}
\begin{figure}[t]
\begin{center}
\includegraphics[width=0.9\columnwidth]{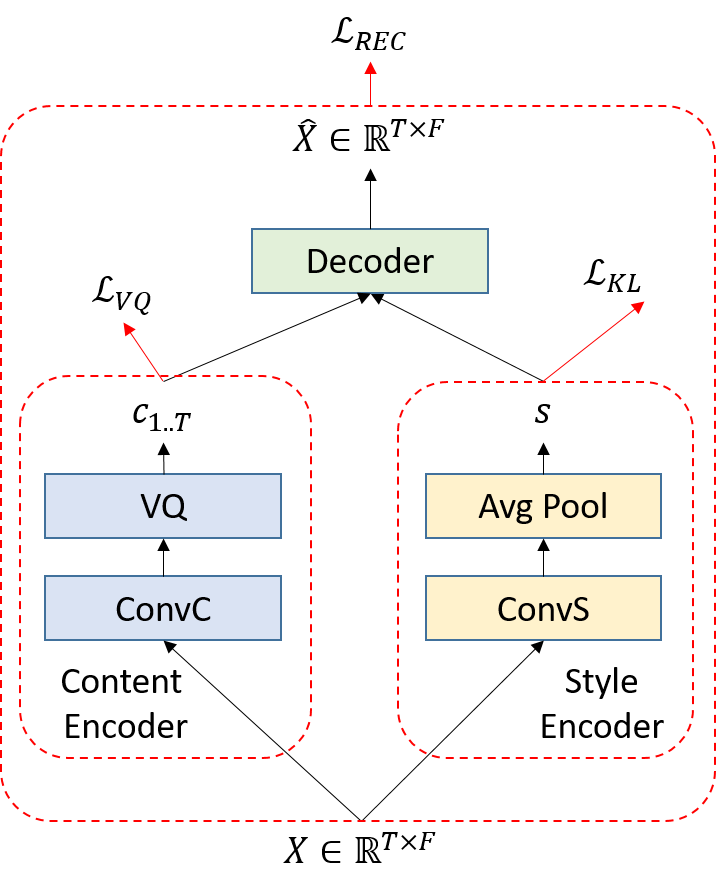}
\caption{Proposed model: disentangling content and style autoencoder.}
\label{fig:model}
\end{center}
\end{figure}
Our proposed model is illustrated in Figure~\ref{fig:model}. On the encoder side, we have two different encoders with different architectures, encoding content and style respectively. On the decoder side, we have a decoder that takes both content and style latent variables to produce speech features.

\subsection{Content Encoder}
\label{sec:content_encoder}
We represent the speech feature as a matrix $X \in \mathbb{R}^{T \times F}$ where $T$ is the frame time-step and $F$ is the feature dimension. 
The content encoder $EncC(\cdot)$ aims to extract a sequence of latent variables $C = [c_1,..,c_T] \in \mathbb{R}^{T \times D_{C}}$ that only represent the content from the input speech $X$. It is built on top of several convolutional layers to represent the content information on the relatively short window time-step.

On the top of the content encoder, we apply information bottleneck via vector quantization (VQ)~\cite{van2017neural} with straight-through gradients. By limiting the number of output possibilities from the content encoder, we force the content encoder and VQ to capture only necessary localized information such as phoneme or subword-like representation and discard their variations (e.g., the speaking style variation). Without using an information bottleneck or strong regularization method such as VQ layer, the content encoder might capture both content and style inside $c_{1..T}$ and the decoder will ignore the style latent variable $s$ from the style encoder.

\subsection{Style Encoder}
The style encoder $EncS(\cdot)$ aims to extract a global latent variable as a vector $s \in \mathbb{R}^{D_{S}}$ that only represent the speaking style from the input speech $X$. It is build on top of several convolutional layers. We summarize the output from the convolutional layer into a single vector with a global average pooling operation across the time-axis. We put a variational layer with Gaussian posterior \cite{Kingma2014} on the top of convolutional layer output. During training, the style variable $s$ is sampled from Eq.~\ref{eq:style}. In the inference step, we directly use $s_\mu$ to represent the style information.

\subsection{Decoder}
The decoder $Dec(.,.)$ takes both content and style latent variables $[c_{1..T}, s]$ and reconstruct the features $\hat{X} \in \mathbb{R}^{T \times F}$ during training stage. For the inference step, we can combine $c_{1..T}$ and $s$ from different input speech and generate a new speech feature that has same content as $C$ and same style as $s$. 

\subsection{Loss}
We optimize our model with multiple loss functions:
\begin{enumerate}
    \item \textbf{Reconstruction loss}: $\mathcal{L}_{REC} = \|X-\hat{X}\|_{1}^{2} + \|X-\hat{X}\|_{2}^{2}.$
    The reconstruction loss minimizes both L1 and L2-norm squared distance between groundtruth $X$ and $\hat{X}$.
    
    \item \textbf{Content VQ loss}:
        \begin{align}
            i &= argmin_{j} \| ConvC(X) - E_j \|_2 \\
            \mathcal{L}_{VQ} &= \| ConvC(X) - sg(E_i) \|_2^{2}
        \end{align} where $sg(\cdot)$ is stop gradient operations.
    The content VQ loss encourages the encoder output $z$ to minimize the distance between itself and the nearest codebook $E_i$.
    
    \item \textbf{Style regularization loss}:
        \begin{align}
            s_{\mu}, s_{\sigma}^2 &= EncS(X); \quad s \sim \mathcal{N}(s_\mu, s_\sigma^2) \label{eq:style} \\ 
            \mathcal{L}_{KL} &= \mathbb{D}_{KL}{(\mathcal{N}(s_\mu, s_\sigma^2) \| \mathcal{N}(0, 1))}
        \end{align}
    The style regularization loss minimizes the KL divergence between the Gaussian posterior $\mathcal{N}(s_\mu, s_\sigma^2)$ with the unit Gaussian prior $\mathcal{N}(0, 1)$.
\end{enumerate}
The final loss is a combined sum:
    \begin{align} \L = \L_{REC} + \gamma \L_{VQ} + \L_{KL} \label{eq:final_loss}.\end{align}

%\rpang{Are there any scaling factors on each loss?}

\subsection{Mutual Information Estimation and Minimization}
Section~\ref{sec:content_encoder} describes the importance of information bottleneck via VQ layer on the content encoder sides. However, the quantized variables $c_{1..T}$ may still manage to carry non-content information such as speaking style.

% There are several possible ways to further decouple content $c_{1..T}$ ad style $s$. First, we may append a speaker classifier on the content encoder output and a gradient reversal layer to flip the gradient sign \cite{meng2018speaker}. However, we need a speaker label from each speech input. Second, 

To further decouple content and style representations without explicit labels, we estimate and minimize the mutual information (MI) from the output between content $c_{1..T}$ with style $s$. By minimizing the mutual information, we reduce the amount of correlation between two random variables \cite{MacKay2003}.

Measuring mutual information directly from high dimensional random variables is a difficult problem~\cite{poole2019variational}. Here, we use InfoNCE~\cite{oord2019representation} to estimate the lower bound MI between content and style:
\begin{align}
    C_i &= AvgPool(ConvC(X_i))\\ \nonumber
    S_i &= AvgPool(ConvS(X_i))\\ \nonumber
    I(C;S) &\geq \mathbb{E}\left[\frac{1}{K}\sum^{K}_{i=1}\log \frac{e^{Sc(C_i,S_i)}}{\frac{1}{K}\sum_{j=1}^{K} e^{Sc(C_i,S_j)}} \right]  \triangleq I_{NCE}. \nonumber
\end{align} where $K$ is the number of samples (across different utterances), $Sc(\cdot, \cdot)$ is a neural network scorer, $C_i$ is the representation from content encoder prior to the quantization, and $S_i$ is the representation from style encoder. The neural network scorer $Sc: \mathbb{R}^{D_c} \times \mathbb{R}^{D_s} \xrightarrow{} \mathbb{R}$ is used to model the density ratio \begin{align} Sc(C_i, S_i) \propto \frac{p(C|S)}{p(C)} \end{align} between content vector $C_i$ and style vector $S_i$. 

Finally, we simultaneously try to maximize $I_{NCE}$ w.r.t $Sc$ and minimize $I_{NCE}$ and $\L$ w.r.t $(\theta=\{EncC,EncS,Dec\})$: \begin{equation}
    \max_{Sc} \min_{EncC, EncS, Dec} (\L + I_{NCE}). \label{eq:loss_final}
\end{equation}
We apply adaptive gradient scaling \cite{pmlr-v80-belghazi18a} to stabilize the loss Eq.~\ref{eq:loss_final} during training.
For each step, the model parameters are updated with:
\begin{align}
    g_{\theta} = \partial &\L / \partial {\theta}; \quad
    g_{a} = \partial I_{NCE} / \partial {\theta}\\
    g_{b} &= min(\|g_{a}\|_2, \|g_{\theta}\|_2) \frac{g_a}{\|g_{a}\|_2} \\
    \theta &\xleftarrow{} \theta - (g_\theta + g_{b}) \\
    Sc &\xleftarrow{} Sc + \partial I_{NCE} / \partial Sc
\end{align}

% \rpang{Explain that mutual information gradient clipping is critical?}

\section{Experiments}
\subsection{Dataset}
To evaluate our proposed model effectiveness on disentangling content and style, we use the Librispeech~\cite{librispeech2015} corpus for all following experiments. To train the generative model, we use all training sets: train-clean 100 hours (contains 251 unique speakers), train-clean 360 hours (contains 921 unique speakers) and train-other 500 hours (contains 1166 unique speakers). We use dev-clean (contains 40 unique speakers) as the development set and test-clean (contains 40 unique speakers) for evaluation. In total, Librispeech contains 2484 unique speakers without overlap between different sets. For the feature extraction, we extract 80 dimensions log Mel-filterbank with window size 25 ms and step size 10 ms. All features are normalized into zero mean and unit standard deviation based on the statistics from the entire dataset. 

\subsection{Hyperparameters}
For the content encoder, we have 10 1D convolutional layers with 768 dimension and residual connections. We use time stride 2 in the 3rd layer. After we fed the input $X$ into the content encoder, the final output length was compressed from $T$ to $T/2$.

For the style encoder, the input $X$ is processed by 6 residual 1D convolution layers with 256 dimension. We set the time stride=2 on three different layers, resulted in 8$\times$ time-length reduction and followed by average pooling operation.

For the decoder, we use 10 1D convolution layers with 768 dimension and residual connection and we feed the style information by concatenating $s$ in the channel axis on \{1,3,5,7\}-th layers. Following \cite{van2017neural}, we set the $\gamma = 0.25$ in the final loss Eq.~\ref{eq:final_loss}.

We evaluate variations of the model along two dimensions.
First, we measure the model performance with different VQVAE codebook sizes with $K \in \{256, 512, 1024\}$. 
Second, we compare the disentanglement performance between models trained with vs. without the mutual information minimization loss. All models are optimized with Adam~\cite{kingma2014adam} optimizer up to 800,000 steps. In total, our model has $\pm$30 million parameters.

% \rpang{Talk about model dim, size, and training steps?}

\subsection{Experiments on Content Preservation}
This set of experiments aim to measure how well the content encoder extracts and preserves the original content from the input speech. There are two different scenarios: 
\begin{enumerate}
    \item No shuffle: feeds speech $X_i$ into both content and style encoder and predict the speech features $\hat{X}$ with the decoder. Since each frame is encoded by 8 to 10 bits of VQVAE codes, this setup tests how well the model compresses speech.
    \item Shuffle: feeds speech $X_i$ into the content encoder and speech $X_j$ ($i \neq j$ ) into the style encoder and predict the speech features $\hat{X}$ with the decoder. This setup measures whether the speech contents are preserved by the VQVAE codes.
\end{enumerate}
We use test-clean for both content and style encoder in order to test the performance of our system with unseen content and speaking style.
To evaluate how similar the predicted speech content compared to the original content from speech $X_i$, we utilize a state-of-the-art pre-trained Conformer L \cite{conformer2020} to transcribe the predicted features $\hat{X}$. After that, we calculate the word error rate for the transcription result with the ground-truth text from input speech $X_i$ on the content encoder side.

Based on Table~\ref{tbl:exp_content}, we observed that increasing the codebook size allows the model to better preserve speech contents, as manifested by the decreasing WER for both shuffle and no shuffle scenarios. Adding the mutual information loss further improves content preservation with decreased WER across all code book sizes, for both shuffle and no-shuffle scenarios.

\begin{table}[h]
\begin{center}
\caption{Predicted features WER between different codebook sizes and with or without mutual information (MI) loss. The Conformer L model's WER on the original test-clean set is $2.1\%$.}
\label{tbl:exp_content}
\begin{tabular}{ |c|c|c|c| } 
 \hline
 \textbf{Codebook} & \textbf{MI} & \multicolumn{2}{c|}{\textbf{WER} ($\%$)}\\
  \textbf{Size} &  \textbf{Loss}  & \textbf{No shuffle} & \textbf{Shuffle} \\
\hline \hline
 256 & No & 4.6 & 7.7\\
     & Yes & 4.4 & 7.0 \\
\hline
 512 & No & 4.2 & 7.0 \\
     & Yes & 3.8 & 6.8 \\
\hline
1024 & No & 3.7 & 6.3 \\
     & Yes & 3.6 & 6.1 \\
 \hline
\end{tabular}
\end{center}
\end{table}

Additionally, we trained a WaveRNN \cite{kalchbrenner2018efficient} model to reconstruct the new speech utterance from different contents and styles without any paired data. The WaveRNN is trained with input content $C_i$ and style $S_i$, extracted from audio $X_i$ with pre-trained content and style encoders. The objective is to minimize the reconstruction loss to the target waveform from audio $X_i$. To synthesize a new speech utterance from WaveRNN that has the same content from utterance $X_i$ and speaking style from utterance $X_j$, we feed the content $C_i$ and style $S_j$ as the input for WaveRNN. We put the generated audio in the following URL: \href{https://unsup-content-style-demo-v1.netlify.app}{https://unsup-content-style-demo-v1.netlify.app}.

\subsection{Experiments on Style Disentanglement}
This set of experiments measure how well the style encoder extracts and preserves the original speaking style from the input speech. We feed different input speech $X_i$ into the content encoder and input speech $X_j$ ($i \neq j$) into the style encoder and predict speech features $\hat{X}$ based on the content and style latent variables.

\begin{figure}[]
\centering
\includegraphics[width=0.99\columnwidth]{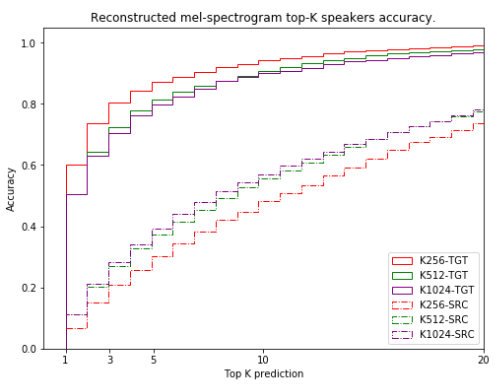}
\caption{Top-K speaker accuracy compared to content (SRC) speaker ID and style (TGT) speaker ID. In this figure, we varied the codebook size and all models are trained with MI loss.
%\rpang{Is the mutual information loss enabled?}
}\label{fig:top_k_speakers}
\end{figure}

Next, we evaluate the speaker similarity between the generated speech and $X_i$ and $X_j$, the input speeches for the content and style encoders, respectively:
\begin{enumerate}
    \item Train a speaker recognition model using ground-truth speaker labels;
    \item Predict the softmax output on the predicted features $\hat{X}$.
\end{enumerate}
Based on the softmax output predictions, we calculate several metrics such as: average ranking, top-1, top-3, and top-5 accuracy compared to the ground truth speaker on $X_j$. As shown in Figure~\ref{fig:top_k_speakers}, the generated speech has a speaker style much closer to $X_j$, the input to the style encoder, than $X_i$, the input to the content encoder.

Table~\ref{tbl:exp_style} further shows how the speaker style accuracy for $X_j$ changes as we vary codebook sizes and whether to enable mutual information loss. We can see that the style and content information becomes more entangled as we increase the codebook size. The mutual information loss helps improving the disentanglement, esp. for the smaller codebooks. Table~\ref{tbl:exp_content} and Table~\ref{tbl:exp_style} together highlight the tradeoffs between content reconstruction and style disentanglement.

\begin{table}[h]
\centering
\caption{Predicted features top-K speaker ranks between different $K$ codebook size and with or without mutual information loss.}
\label{tbl:exp_style}
\begin{tabular}{ |c|c|c|c|c|c| } 
 \hline
 \textbf{Codebook} & \textbf{MI} & \textbf{Avg}. & \textbf{Top-1} & \textbf{Top-3} & \textbf{Top-5} \\
\hline \hline
 256 & No & 3.9 & 0.5 & 0.72 & 0.81 \\
     & Yes & 2.9 & 0.6 & 0.8 & 0.87 \\
\hline
 512 & No & 3.9 & 0.5 & 0.71 & 0.8 \\
     & Yes & 3.7 & 0.5 & 0.73 & 0.81 \\
\hline
1024 & No & 4.0 & 0.52 & 0.72 & 0.8 \\
     & Yes & 4.0  & 0.51 & 0.71 & 0.8 \\
 \hline
\end{tabular}
\end{table}

\subsection{Experiments on Few-Shot Speaker Recognition}
\begin{figure}[h]

\centering
\includegraphics[width=0.5\columnwidth]{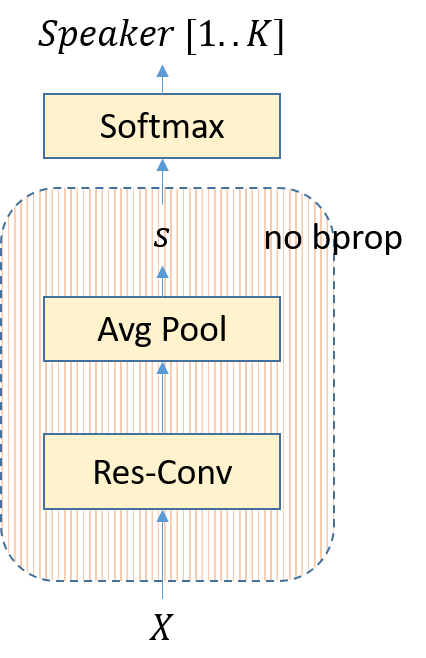}
\caption{Using pre-trained style encoder as a speaker recognition by appending and fine-tuning a softmax layer.} \label{fig:few_shot_style}

\end{figure}

Recently, there are many success stories for pre-training a model with a large amount of unlabeled data and fine-tune it with a small labeled data \cite{schneider2019wav2vec, devlin2018bert}. In this experiment, we fine-tune a pre-trained style encoder for the few-shot speaker recognition task. For the dataset, we split the train-dev-test from train set 1000h (with a total unique speaker 2338) by sampling 1 or 3 utterance per speaker for the train, 5\% for the dev and 5\% for the test without overlap.
As shown in Figure~\ref{fig:few_shot_style}, we add a linear projection layer on the top of the style encoder to compute the logits for speaker classification. During the fine-tuning, we freeze all parameters except the final softmax projection layer. 

We consider three scenarios for fine-tuning a speaker recognition model: 1) train from scratch without pre-trained weight (without freezing any weight), 2) initialize with a style encoder pre-trained without MI loss (freeze style encoder weight), 3) initialize with a style encoder pre-trained with MI loss (freeze style encoder weight). We also vary the amount of fine-tuning data with 1 or 3 examples per speaker, termed as ``1-shot'' and ``3-shot'', respectively, in Table~\ref{tbl:few_shot_result}. The results show that we can achieve $>90\%$ accuracy with 1-shot fine-tuning, compared with $<11\%$ without a pre-trained encoder, and $>95\%$ accuracy with 3 examples per speaker, compared with $63\%$ without pre-training. Further, the use of mutual information loss brings small improvements to the results.

\begin{table}[h]
\centering
\caption{1-shot and 3-shot accuracy based on three difference scenarios. Model M1: model trained from scratch, M2: pre-trained style encoder without MI loss, M3: pre-trained style encoder with MI loss.}
\label{tbl:few_shot_result}
\begin{tabular}{ |c|c|c| } 
 \hline
 \textbf{Model} & \textbf{1-shot acc} ($\%$)& \textbf{3-shot acc} ($\%$)\\
\hline \hline
M1 & 10.8 & 63.0 \\
\hline
M2 & 90.5 & 95.9 \\
\hline
M3 & 91.4 & 96.4 \\
\hline
\end{tabular}
\end{table}

\section{Conclusion}

We present a framework to learn disentangled representations of speech with unlabeled data. The framework includes a local VQ encoder to extract a discrete sequence representation of the speech contents and a global VAE encoder to learn a continuous representation of speech styles. Our evaluation shows that the discrete sequence representation effectively captures the linguistic contents while the continuous global representation encapsulates the speaker style. Additionally, we also show the application from our pre-trained model, in which we successfully train a speaker recognition system with high accuracy ($>90\%$) only with one sample per speaker.

% \vfill\pagebreak

\bibliographystyle{IEEEtran}

\bibliography{refs}

\end{document}